\pdfoutput=1
\documentclass[11pt,letterpaper]{article}
\usepackage[T1]{fontenc}
\usepackage{lmodern}
\usepackage[margin=1in]{geometry}
\usepackage{microtype}
\usepackage{amsmath}
\usepackage{array,longtable,booktabs,calc}
\usepackage{etoolbox}
\usepackage{fvextra}
\usepackage{tikz}
\usepackage{xurl}
\usepackage[hidelinks,pdftitle={The Noise Floor of a Quantization Comparison},pdfauthor={Aleks Blagojevich, Trevor Sunderland}]{hyperref}
\newcommand{\tablesize}{\footnotesize}
\AtBeginEnvironment{longtable}{\tablesize}
\providecommand{\tightlist}{\setlength{\itemsep}{0pt}\setlength{\parskip}{0pt}}
\definecolor{floorc}{HTML}{2A78D6}
\definecolor{quantc}{HTML}{EB6834}
\definecolor{gridc}{HTML}{E4E3DF}
\definecolor{inkc}{HTML}{52514E}
\title{The Noise Floor of a Quantization Comparison\\[0.4em]\large Engine, restart, batch and kernel drift, measured against 8-bit and 4-bit effects in two decision models}
\author{Aleks Blagojevich\\\footnotesize Level5\\\footnotesize\texttt{aleks@level5.pro} \and Trevor Sunderland\\\footnotesize Level5\\\footnotesize\texttt{trevor@level5.pro}}
\date{\footnotesize General contact: \texttt{hello@level5.pro}}
\begin{document}
\maketitle
\begin{abstract}
A quantized model is usually judged by comparing it with its BF16 original, but that comparison changes more than the weights. It often changes the serving engine, it always runs on a different server instance, it may change how requests are batched, and on some stacks the library selects a different kernel depending only on which packages import. We measure each of these sources on the same 2,400 public decisions for two of Cloudflare\textquotesingle s decision models, Clef-flash (9B) and Clef (27B), on an NVIDIA GB10, and state every quantization effect as the excess over them. Sources that are not quantization move between 0 and 0.92\% of decisions. llama.cpp restarts are bit-identical on every record for both models. vLLM restarts flip 0, 5 and 20 decisions at BF16, FP8 and NVFP4 on Clef-flash and 4, 0 and 22 on Clef; a third instance of Clef\textquotesingle s quantized builds (exploratory) leaves FP8 at 0 and puts NVFP4 at 4 to 22, depending on which two instances are compared. An engine change at BF16 flips 6 to 16, a batch-size change 8 to 15, and the Gated DeltaNet kernel transformers selects unasked 10 to 14. Quantization inside one engine moves more: for Clef-flash and Clef, 0.88\% and 1.04\% at Q8\_0, 2.12\% and 1.62\% at FP8, 5.79\% and 3.50\% at Q4\_K\_M, and 5.79\% and 5.67\% at NVFP4. On both models, every 4-bit interval excludes every non-quantization source. Two of our five pre-registered predictions failed, and the same two fail when the tests are rerun on the 27B: FP8 is not inside restart noise, and changing engine moves fewer decisions than 8-bit quantization does. Accuracy hides most of this. Over the full 31,275 public questions, Clef\textquotesingle s Q4\_K\_M build moves accuracy by a net six questions (-0.02 points {[}-0.20, +0.16{]}) while 436 of its decisions get worse (exploratory), and NVFP4 costs Clef-flash 0.95 points and Clef 0.65. We release the code, quantization recipes, per-record outputs and the pre-registration.
\end{abstract}
\section{1. Introduction}

Decision models answer a fixed question with one option from a fixed set: which of 77 intents, whether a premise entails a hypothesis, whether an agent should call a tool. Their output is a probability distribution over those options, and the deployed behavior is the top option. A change in the top option is a changed business decision. The quantity that matters to an operator is therefore how many decisions a build changes, not only how many it gets right. Dutta et al. (2024) make the same argument for language models in general and use flips as the metric.

Accuracy hides most of this. Over 31,275 public questions, a net change of six questions hid 436 that got worse: that is the 4-bit GGUF build of the 27B model we study, against its BF16 original. On the 9B model, the same build\textquotesingle s net change of 114 questions hid 527 that got worse. This split is exploratory, and the full-set results give it for every build.

Quantization evaluations rarely isolate the weights. A typical comparison runs the quantized build once on a serving engine (vLLM for FP8 and NVFP4, llama.cpp for GGUF) and the reference once on another, and attributes any difference to quantization. At least four other things changed:

\begin{enumerate}
\def\labelenumi{\arabic{enumi}.}
\tightlist
\item
  \textbf{The engine.} transformers, vLLM and llama.cpp implement the same architecture with different kernels and accumulation orders.
\item
  \textbf{The server instance.} We measured on 2026-10-16 that one vLLM FP8 instance reproduces itself bit for bit, while a fresh instance with identical flags differs by up to 0.94 in a logit.
\item
  \textbf{The batch.} vLLM is not batch-invariant: the logits of one request depend on what it was batched with.
\item
  \textbf{The kernel.} transformers 5.x routes the models\textquotesingle{} Gated DeltaNet layers through a decorator that tries three implementations in order: a hub kernel, the flash-linear-attention (\texttt{fla}) package and a pure-PyTorch fallback. Which one runs depends only on whether a package imports, and in the version we used nothing reports which ran.
\end{enumerate}

We keep two claims apart throughout. A quantized build that answers differently from its BF16 original was changed on purpose: its flips are the price of the memory it saves, and the question is how large that price is. A server instance that answers differently from an identically configured instance of the same build was not changed at all. That is nondeterminism, a defect to fix rather than a price to pay. We report both and never pool them.

We measure each source on the same records and report each quantization effect as the paired difference against them. Our contributions:

\begin{itemize}
\tightlist
\item
  A decomposition protocol for quantization comparisons, built on paired flip-rate differences with bootstrap intervals on a fixed record set.
\item
  Measurements of three engines and five weight formats for a 9B and a 27B decision model on one Blackwell-class device, with every record\textquotesingle s output released.
\item
  Five pre-registered predictions with their outcomes, including the two that failed, and an exploratory rerun of all five on the second model.
\item
  An exploratory split of every full-set flip into decisions that got worse, got better, or changed from one wrong answer to another, and each quantization effect stated as a multiple of the largest non-quantization source.
\item
  An observation we did not anticipate: the vLLM restart floor depends on the format, but not in a fixed order. Clef-flash\textquotesingle s floor rises from BF16 to FP8 to NVFP4. Clef\textquotesingle s FP8 restart is the quietest of its three, and its NVFP4 restart is the noisiest non-quantization source we measured on either model. llama.cpp\textquotesingle s single-sequence decision server has a restart floor of exactly zero on both. A third instance shows that the vLLM floor also depends on which two instances are compared, while the quantization effect measured against the reference stays within two flips across all three.
\end{itemize}

\section{2. Related work}

\textbf{Flips as the metric.} Dutta et al. (2024, "Accuracy is Not All You Need", \href{https://arxiv.org/abs/2407.09141}{arXiv:2407.09141}) show that compressed models can match the baseline\textquotesingle s accuracy while changing many individual answers, and argue for reporting flips. We adopt flips as the primary metric and add the question of how many flips a comparison would show with no quantization at all. Wu, Dhiman and Koshiyama (2026, \href{https://arxiv.org/abs/2608.06564}{arXiv:2608.06564}) explain why the flips happen: quantization shrinks decision margins roughly linearly, and the fitted shrinkage predicts flip rates across 16 models to a median of 1.0 point. Our finding that flips concentrate where the reference is unsure is what their account predicts for a decision head.

\textbf{Backend variability.} Masoudian et al. (2026, \href{https://arxiv.org/abs/2608.04714}{arXiv:2608.04714}) attribute roughly 39\% of the benchmark variability a practitioner sees out of the box to the inference backend. Pape, Evertz and Sch\"onherr (2026, \href{https://arxiv.org/abs/2605.19537}{arXiv:2605.19537}) find that the backend alone moves benchmark scores by up to 16.6 points across five engines, vLLM and llama.cpp among them. Both study generation, and neither studies quantization or server restarts. We measure the backend effect at BF16 as one source among four and set it directly against quantization on the same records.

\textbf{Nondeterminism within a server.} Thinking Machines Lab (He, 2025, "Defeating Nondeterminism in LLM Inference") trace run-to-run differences inside one server mainly to batch variance and release batch-invariant kernels. Yuan et al. (2025, \href{https://arxiv.org/abs/2506.09501}{arXiv:2506.09501}) report accuracy swings of up to 9\% from GPU count, GPU type and batch size. Li et al. (2026, \href{https://arxiv.org/abs/2610.42148}{arXiv:2610.42148}) find that concurrency alone changes some answers of Qwen3.5 decision workloads in every engine they test, up to 3.2\% in vLLM between one and 64 clients. Our restart measurement is a different quantity: two instances started with identical flags.

\textbf{Serving context in classifiers.} Kasa, Kasa and Negi (2026, \href{https://arxiv.org/abs/2610.09111}{arXiv:2610.09111}) is the closest work that counts changed decisions. They train 180 text classifiers and measure label changes under changes of batch shape, precision, device and 8-bit quantization against a repeatability floor, and they find the changes concentrated at small margins. They leave production engines out of scope, noting that vLLM and SGLang "require their own measurement", and they do not study restarts. That is the case we measure, on two released decision models.

\textbf{Restarts, kernel swaps and quantization.} Chen (2026, \href{https://arxiv.org/abs/2608.13756}{arXiv:2608.13756}) swaps only the INT8 matrix-multiply kernel inside vLLM: each kernel reproduced itself bit for bit across cold restarts at 1.7B, and the two agreed on no generated sequence. Patodiya (2026, \href{https://arxiv.org/abs/2609.04748}{arXiv:2609.04748}) shows that prefix-cache reuse makes repeated llama.cpp and vLLM runs diverge, and that the divergence grows with quantization, from 37.4\% of agentic episodes at F16 to 82.5\% at Q4\_K\_M. Without prefix reuse every run repeated exactly. Our vLLM servers run with prefix caching off, so that mechanism is not behind our floor. A vLLM bug report (vllm-project/vllm\#58899, 2026-09-27) shows greedy output changing between fresh processes because torch.compile picks a fused kernel\textquotesingle s reduction configuration by timing it at first launch, and processes that picked the same configuration were bit-identical. A second report (vllm-project/vllm\#62820, 2026-10-21) shows FlashInfer\textquotesingle s FP4 GEMM autotuning changing temperature-0 tokens of a 27B NVFP4 model. We test below whether start-up selection of this kind explains our restart floor. Kozyrev and Maiboroda (2026, \href{https://arxiv.org/abs/2609.04098}{arXiv:2609.04098}) quantize all of Qwen3.8-27B to NVFP4, Gated DeltaNet gate projections included, and find the gates the least sensitive layers. Our builds keep those projections in BF16, which by their account is conservative.

\textbf{Community measurements of these models.} Hugging Face model cards already report agreement between quantized Clef builds and BF16. TurinTech (2026) measures ggml-org\textquotesingle s 27B Q8\_0 and Q4\_K\_M on 2,000 typed decisions on three GPUs, a GB10 among them, at 98.8 to 99.0\% and 96.9\% same answer. kurcontko (2026) measures the 9B at FP8 and NVFP4 in vLLM on 1,400 records (98.8\% and 94.3\% agreement) and vLLM BF16 against transformers (99.8\%). Our full-set rates agree with theirs: the 27B Q4\_K\_M flips 3.20\% of 31,275 public questions, which is 96.8\% agreement. None of these cards sets a quantization effect against a restart or kernel floor.

\textbf{A quantization effect against an engine floor.} Lu and Ni (2026, \href{https://arxiv.org/abs/2611.12634}{arXiv:2611.12634}) make the comparison closest to ours. They predict the KL divergence that int8, fp8 and int4 quantization cause in five generative language models, and they set it against a vLLM noise floor of 0.17 to 0.22 nats per sequence, measured by changing the reduction order: other kernel selections, tensor parallelism, batch-invariant kernels. Mismatch, they write, "is only meaningful above the run-to-run noise floor of the serving engine." That is the comparison we make, in a different currency and from a different floor. Their reruns that reused the kernel selection were bitwise identical, so their floor comes from deliberate changes. Ours comes from restarting servers with identical flags, in two engines, measured in decision flips on decision models, with a kernel switch and a batch change beside it.

We searched on 2026-11-03 (arXiv, 20 keyword queries; the web; the model cards of all 191 Clef derivatives on Hugging Face, with silent kernel selection searched specifically) and searched again on 2026-11-10 (about 470 arXiv abstracts submitted or updated since 2026-10-25; vLLM, transformers and flash-linear-attention issues; the 60 most recently modified Clef models). We found no paper that measures the restart-to-restart decision flips of a production serving engine and sets a quantization effect against them on the same records. Kernel choice as a source of numerical difference is well established (Chen; Kasa et al.; Lu and Ni; vllm\#58899). We did not find our kernel-selection case measured anywhere: a library that picks an implementation according to which packages import. Since version 5.17.0, transformers warns once when the PyTorch fallback runs (pull request \#48185). Version 5.16.1, which we used, reports nothing, and the warning does not cover the choice between the hub kernel and \texttt{fla}.

\section{3. Setup}

\subsection{3.1 Models, hardware and software}

Clef-flash (9B) and Clef (27B) are Cloudflare\textquotesingle s released decision models (Cloudflare, 2026). Each adds a joint schema head that scores a question\textquotesingle s options. Both are built on the Qwen3.5 hybrid architecture, with a full-attention layer every fourth layer and Gated DeltaNet layers (Yang et al., 2024) in between: Clef-flash has 24 Gated DeltaNet layers of 32, and Clef 48 of 64 (from the models\textquotesingle{} configuration files). All runs used one NVIDIA DGX Spark (GB10, 128 GB unified memory) inside NVIDIA\textquotesingle s NGC vLLM 26.09 container (CUDA 13.4 forward-compatibility mode).

{\def\LTcaptype{none}
\begin{longtable}[]{@{}
  >{\raggedright\arraybackslash}p{(\linewidth - 6\tabcolsep) * \real{0.3100}}
  >{\raggedright\arraybackslash}p{(\linewidth - 6\tabcolsep) * \real{0.1700}}
  >{\raggedright\arraybackslash}p{(\linewidth - 6\tabcolsep) * \real{0.2400}}
  >{\raggedright\arraybackslash}p{(\linewidth - 6\tabcolsep) * \real{0.2800}}@{}}
\toprule\noalign{}
\begin{minipage}[b]{\linewidth}\raggedright
Engine
\end{minipage} & \begin{minipage}[b]{\linewidth}\raggedright
Version
\end{minipage} & \begin{minipage}[b]{\linewidth}\raggedright
Used for
\end{minipage} & \begin{minipage}[b]{\linewidth}\raggedright
Batching during quality runs
\end{minipage} \\
\midrule\noalign{}
\endhead
\bottomrule\noalign{}
\endlastfoot
transformers (Cloudflare\textquotesingle s \texttt{load\_release\_model}) & 5.16.1 & \texttt{ref}, \texttt{ref-r2}, \texttt{ref-fla} & batch 1 (\texttt{ref}, \texttt{ref-fla}); batch 8 (\texttt{ref-r2}) \\
vLLM (Kwon et al., 2023) with a Clef decision plugin & 0.29.0+5013de39 & \texttt{bf16-srv}, \texttt{fp8}, \texttt{nvfp4} and their restarts & max-num-seqs 16, 16 requests in flight \\
llama.cpp decision server & b11671 (ba3c05f7, includes PR \#29831) & \texttt{gguf-bf16}, \texttt{q8\_0}, \texttt{q4\_k\_m} and the BF16 restart & single sequence \\
\end{longtable}
}

\subsection{3.2 Configurations}

Both models ran the same 13 configurations. Clef also ran two exploratory restart probes after the sweep.

{\def\LTcaptype{none}
\begin{longtable}[]{@{}
  >{\raggedright\arraybackslash}p{(\linewidth - 6\tabcolsep) * \real{0.3600}}
  >{\raggedright\arraybackslash}p{(\linewidth - 6\tabcolsep) * \real{0.2400}}
  >{\raggedright\arraybackslash}p{(\linewidth - 6\tabcolsep) * \real{0.1600}}
  >{\raggedright\arraybackslash}p{(\linewidth - 6\tabcolsep) * \real{0.2400}}@{}}
\toprule\noalign{}
\begin{minipage}[b]{\linewidth}\raggedright
id
\end{minipage} & \begin{minipage}[b]{\linewidth}\raggedright
Format
\end{minipage} & \begin{minipage}[b]{\linewidth}\raggedright
Engine
\end{minipage} & \begin{minipage}[b]{\linewidth}\raggedright
Role
\end{minipage} \\
\midrule\noalign{}
\endhead
\bottomrule\noalign{}
\endlastfoot
\texttt{ref} & BF16 & transformers, PyTorch GDN fallback & ground truth \\
\texttt{bf16-srv} & BF16 & vLLM & engine baseline for FP8 and NVFP4 \\
\texttt{fp8} & FP8 W8A8 (llm-compressor) & vLLM & 8-bit, Blackwell-native \\
\texttt{nvfp4} & NVFP4 W4A4 (llm-compressor) & vLLM & 4-bit, Blackwell-native \\
\texttt{gguf-bf16} & GGUF BF16 & llama.cpp & engine baseline for GGUF \\
\texttt{q8\_0} & GGUF Q8\_0 & llama.cpp & the 8-bit most local users run \\
\texttt{q4\_k\_m} & GGUF Q4\_K\_M & llama.cpp & the 4-bit most local users run \\
\texttt{bf16-srv-r2}, \texttt{fp8-r2}, \texttt{nvfp4-r2}, \texttt{gguf-bf16-r2} & as base & as base & new server instance, same flags \\
\texttt{fp8-r3}, \texttt{nvfp4-r3} (Clef only, 2026-11-03, exploratory) & as base & as base & third server instance, same flags \\
\texttt{ref-r2} & BF16 & transformers, batch 8 & rerun with a batch-size change \\
\texttt{ref-fla} & BF16 & transformers, \texttt{fla} 0.5.2 GDN kernel & kernel selection \\
\end{longtable}
}

File sizes in GB:

{\def\LTcaptype{none}
\begin{longtable}[]{@{}lllllll@{}}
\toprule\noalign{}
Model & \texttt{ref} & \texttt{gguf-bf16} & \texttt{fp8} & \texttt{q8\_0} & \texttt{nvfp4} & \texttt{q4\_k\_m} \\
\midrule\noalign{}
\endhead
\bottomrule\noalign{}
\endlastfoot
Clef-flash & 19.06 & 18.16 & 12.15 & 9.66 & 9.13 & 5.72 \\
Clef & 54.97 & 54.06 & 30.65 & 28.73 & 20.00 & 16.59 \\
\end{longtable}
}

Every quantized build keeps the joint schema head, the vision tower, the norms and the Gated DeltaNet gate projections in BF16. Calibration data for FP8 and NVFP4 comes only from the tasks\textquotesingle{} training splits, so it never overlaps a scored record. We release the recipes, and the calibration sets follow the same record-or-ID rule as the scored records (see Code and data availability).

\subsection{3.3 Tasks}

Twelve public tasks with fixed, seeded held-out splits, 31,275 questions in total:

\begin{itemize}
\tightlist
\item
  intent: BANKING77 (3,080; Casanueva et al., 2020) and CLINC150 with its out-of-scope class (5,000; Larson et al., 2019)
\item
  inference: ANLI R1 (1,000), R2 (1,000) and R3 (1,200) (Nie et al., 2019), and ContractNLI (1,991; Koreeda and Manning, 2021)
\item
  knowledge: MMLU (5,000; Hendrycks et al., 2020), ARC-Challenge (1,172; P. Clark et al., 2018) and BoolQ (3,270; C. Clark et al., 2019)
\item
  grounding: RAGTruth (2,700; Niu et al., 2024)
\item
  agent: When2Call (3,652; Ross et al., 2025)
\item
  ordinal: SST-5 (2,210; Socher et al., 2013)
\end{itemize}

Each record is posed in Clef\textquotesingle s own question schema (choice, yes/no, or ordinal score) with one fixed prompt format, released with the results. We also scored the authors\textquotesingle{} private operational data: four sets, 13,415 records, 27,738 questions. It plays no part in this report\textquotesingle s argument, because nobody else can reproduce it. We report it only as aggregate numbers.

\subsection{3.4 Comparison unit and statistics}

The restart repeats and \texttt{ref-fla} ran on the first 200 records of each public task. Every decomposition comparison therefore uses exactly those 2,400 records for every configuration of a model. Full-set numbers appear in the full-set section and the appendix.

The two units are not interchangeable, so every rate below names its unit. The 2,400 records are the first 200 of each task, not a random sample, and they weight every task equally, where the full set weights tasks by size. Weighting tasks equally alone raises each quantized build\textquotesingle s full-set flip rate against \texttt{ref} by 1.26 to 1.36 times. On Clef-flash the 2,400-record rates against \texttt{ref} run 1.3 to 1.6 times the full-set rates. On Clef they run 0.8 to 1.2 times, because which records come first moves the rate in both directions (exploratory; \nolinkurl{analysis/exploratory_checks.md}).

For each pair of configurations we report:

\begin{itemize}
\tightlist
\item
  the top-1 flip rate
\item
  mean total-variation distance (TVD) between the option distributions
\item
  maximum absolute probability difference
\item
  the share of records that are bit-identical. That means all logits are equal or, for llama.cpp, all probabilities are equal, because llama.cpp returns full-precision probabilities and no logits.
\end{itemize}

Intervals are percentile 95\% intervals from 2,000 paired bootstrap resamples over records with a fixed seed (20261015). One resample matrix is shared across all pairs of a model, so differences between pairs are themselves paired. Where a pair has zero flips the bootstrap interval collapses, and we give the exact Clopper-Pearson interval (Clopper and Pearson, 1934) instead: 0 of 2,400 bounds the rate above at 0.15\%.

\subsection{3.5 Making the kernel observable}

Because transformers chooses the Gated DeltaNet kernel silently, we made the choice explicit. In \texttt{torch} mode the worker blocks the \texttt{fla} import for its own process, which forces the fallback even if the package is installed. In \texttt{fla} mode it wraps \texttt{fla}\textquotesingle s entry points with counters and asserts after warm-up that they actually ran. Every result line records the kernel used. Every 27B result line carries this tag.

Six early Clef-flash \texttt{ref} sets predate the tag. Three independent pieces of evidence show they ran the fallback:

\begin{itemize}
\tightlist
\item
  a later \texttt{torch}-mode run reproduced the sweep\textquotesingle s stored \texttt{ref} output for ANLI R1 exactly (maximum logit difference 0.0 on 200 questions)
\item
  \texttt{fla} was never present on the sweep container\textquotesingle s library path
\item
  the first \texttt{ref-fla} run failed precisely because \texttt{fla} could not be imported in that container
\end{itemize}

\section{4. Results}

\subsection{4.1 The decomposition}

All pairs are on the same 2,400 decisions per model. Each cell gives the number of flips, the flip rate {[}95\% CI{]} and the share of bit-identical records. The Clef-flash column holds the pre-registered measurement. The Clef column reruns it on the second model and is exploratory.

{\def\LTcaptype{none}
\begin{longtable}[]{@{}
  >{\raggedright\arraybackslash}p{(\linewidth - 6\tabcolsep) * \real{0.2500}}
  >{\raggedright\arraybackslash}p{(\linewidth - 6\tabcolsep) * \real{0.1900}}
  >{\raggedright\arraybackslash}p{(\linewidth - 6\tabcolsep) * \real{0.2800}}
  >{\raggedright\arraybackslash}p{(\linewidth - 6\tabcolsep) * \real{0.2800}}@{}}
\toprule\noalign{}
\begin{minipage}[b]{\linewidth}\raggedright
Source
\end{minipage} & \begin{minipage}[b]{\linewidth}\raggedright
Pair
\end{minipage} & \begin{minipage}[b]{\linewidth}\raggedright
Clef-flash (9B)
\end{minipage} & \begin{minipage}[b]{\linewidth}\raggedright
Clef (27B)
\end{minipage} \\
\midrule\noalign{}
\endhead
\bottomrule\noalign{}
\endlastfoot
\multicolumn{4}{@{}>{\raggedright\arraybackslash}p{(\linewidth - 6\tabcolsep) * \real{1.0000} + 6\tabcolsep}@{}}{
\textbf{Not quantization}} \\
Restart, llama.cpp BF16 & \texttt{gguf-bf16} vs \texttt{-r2} & 0, 0.00\% {[}exact 0.00, 0.15{]}, 100.0\% & 0, 0.00\% {[}exact 0.00, 0.15{]}, 100.0\% \\
Restart, vLLM BF16 & \texttt{bf16-srv} vs \texttt{-r2} & 0, 0.00\% {[}exact 0.00, 0.15{]}, 93.7\% & 4, 0.17\% {[}0.04, 0.33{]}, 6.7\% \\
Restart, vLLM FP8 & \texttt{fp8} vs \texttt{-r2} & 5, 0.21\% {[}0.04, 0.42{]}, 15.0\% & 0, 0.00\% {[}exact 0.00, 0.15{]}, 95.3\% \\
Restart, vLLM NVFP4 & \texttt{nvfp4} vs \texttt{-r2} & 20, 0.83\% {[}0.50, 1.21{]}, 76.8\% & 22, 0.92\% {[}0.54, 1.33{]}, 87.0\% \\
Engine at BF16, transformers to llama.cpp & \texttt{ref} vs \texttt{gguf-bf16} & 6, 0.25\% {[}0.08, 0.46{]}, 0.0\% & 16, 0.67\% {[}0.33, 1.00{]}, 0.0\% \\
Engine at BF16, transformers to vLLM & \texttt{ref} vs \texttt{bf16-srv} & 8, 0.33\% {[}0.12, 0.58{]}, 6.3\% & 13, 0.54\% {[}0.25, 0.83{]}, 2.1\% \\
Batch size 1 to 8, native & \texttt{ref} vs \texttt{ref-r2} & 8, 0.33\% {[}0.12, 0.58{]}, 7.2\% & 15, 0.62\% {[}0.33, 0.96{]}, 2.6\% \\
GDN kernel, fallback to \texttt{fla} & \texttt{ref} vs \texttt{ref-fla} & 10, 0.42\% {[}0.21, 0.71{]}, 6.8\% & 14, 0.58\% {[}0.29, 0.88{]}, 2.0\% \\
\multicolumn{4}{@{}>{\raggedright\arraybackslash}p{(\linewidth - 6\tabcolsep) * \real{1.0000} + 6\tabcolsep}@{}}{
\textbf{Quantization, against BF16 in the same engine}} \\
Q8\_0, llama.cpp & \texttt{gguf-bf16} vs \texttt{q8\_0} & 21, 0.88\% {[}0.54, 1.29{]}, 0.0\% & 25, 1.04\% {[}0.67, 1.46{]}, 0.0\% \\
FP8, vLLM & \texttt{bf16-srv} vs \texttt{fp8} & 51, 2.12\% {[}1.54, 2.71{]}, 0.1\% & 39, 1.62\% {[}1.12, 2.12{]}, 0.0\% \\
Q4\_K\_M, llama.cpp & \texttt{gguf-bf16} vs \texttt{q4\_k\_m} & 139, 5.79\% {[}4.92, 6.79{]}, 0.0\% & 84, 3.50\% {[}2.79, 4.29{]}, 0.0\% \\
NVFP4, vLLM & \texttt{bf16-srv} vs \texttt{nvfp4} & 139, 5.79\% {[}4.88, 6.71{]}, 0.0\% & 136, 5.67\% {[}4.79, 6.58{]}, 0.0\% \\
\end{longtable}
}

Mean TVD for every pair is in \nolinkurl{analysis/decomposition.md} (Clef-flash) and \nolinkurl{analysis/clef_decomp/decomposition.md} (Clef).

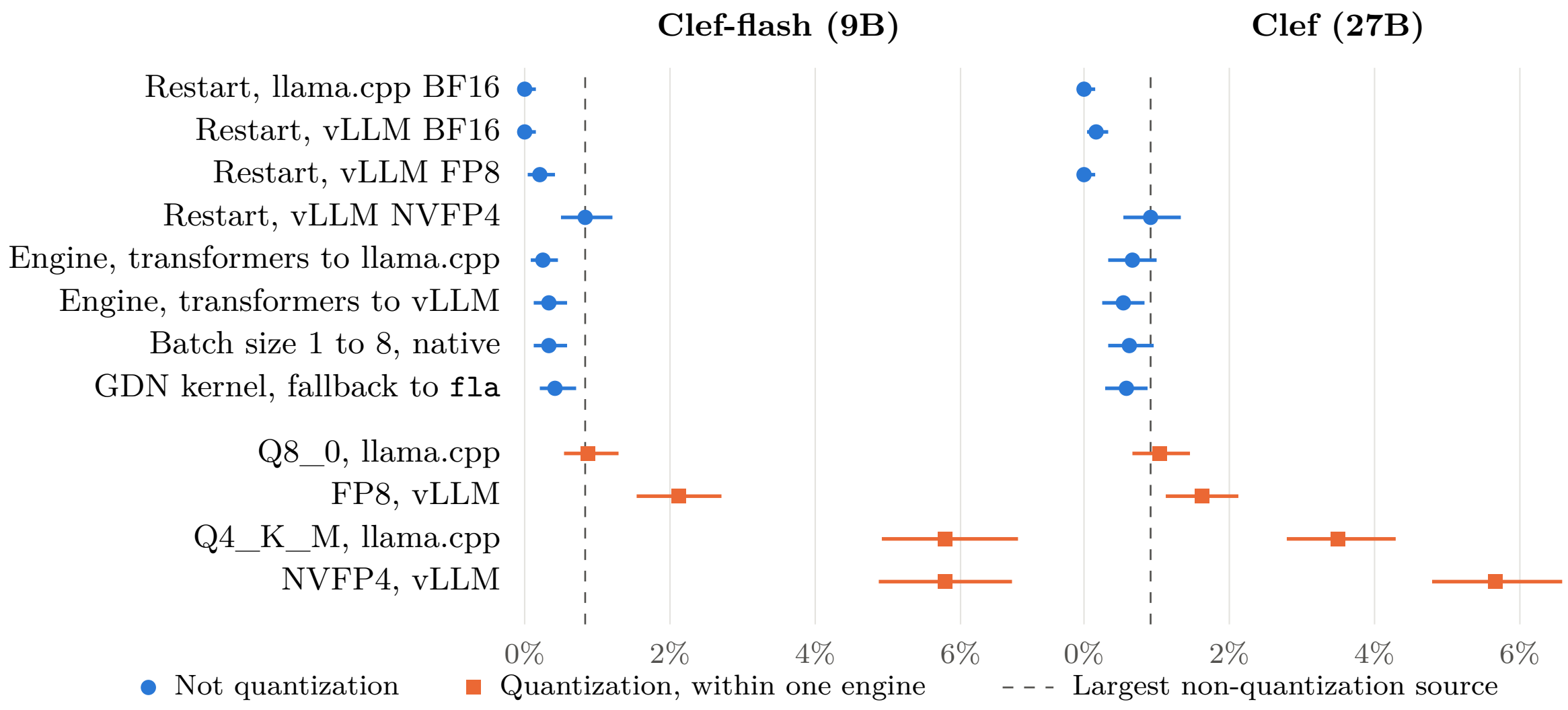
\begin{figure}[htbp]
\centering
\begin{tikzpicture}[x=1cm,y=1cm]
\node[font=\footnotesize,anchor=east] at (4.800,-0.210) {Restart, llama.cpp BF16};
\node[font=\footnotesize,anchor=east] at (4.800,-0.630) {Restart, vLLM BF16};
\node[font=\footnotesize,anchor=east] at (4.800,-1.050) {Restart, vLLM FP8};
\node[font=\footnotesize,anchor=east] at (4.800,-1.470) {Restart, vLLM NVFP4};
\node[font=\footnotesize,anchor=east] at (4.800,-1.890) {Engine, transformers to llama.cpp};
\node[font=\footnotesize,anchor=east] at (4.800,-2.310) {Engine, transformers to vLLM};
\node[font=\footnotesize,anchor=east] at (4.800,-2.730) {Batch size 1 to 8, native};
\node[font=\footnotesize,anchor=east] at (4.800,-3.150) {GDN kernel, fallback to \texttt{fla}};
\node[font=\footnotesize,anchor=east] at (4.800,-3.790) {Q8\_0, llama.cpp};
\node[font=\footnotesize,anchor=east] at (4.800,-4.210) {FP8, vLLM};
\node[font=\footnotesize,anchor=east] at (4.800,-4.630) {Q4\_K\_M, llama.cpp};
\node[font=\footnotesize,anchor=east] at (4.800,-5.050) {NVFP4, vLLM};
\node[font=\footnotesize\bfseries,anchor=south] at (7.400,0.100) {Clef-flash (9B)};
\draw[gridc,line width=0.4pt] (4.900,-5.470) -- (4.900,0.000);
\node[font=\scriptsize,anchor=north,text=inkc] at (4.900,-5.500) {0\%};
\draw[gridc,line width=0.4pt] (6.329,-5.470) -- (6.329,0.000);
\node[font=\scriptsize,anchor=north,text=inkc] at (6.329,-5.500) {2\%};
\draw[gridc,line width=0.4pt] (7.757,-5.470) -- (7.757,0.000);
\node[font=\scriptsize,anchor=north,text=inkc] at (7.757,-5.500) {4\%};
\draw[gridc,line width=0.4pt] (9.186,-5.470) -- (9.186,0.000);
\node[font=\scriptsize,anchor=north,text=inkc] at (9.186,-5.500) {6\%};
\draw[inkc,dashed,line width=0.5pt] (5.495,-5.470) -- (5.495,0.000);
\draw[floorc,line width=1pt] (4.900,-0.210) -- (5.010,-0.210);
\fill[floorc] (4.900,-0.210) circle (0.075);
\draw[floorc,line width=1pt] (4.900,-0.630) -- (5.010,-0.630);
\fill[floorc] (4.900,-0.630) circle (0.075);
\draw[floorc,line width=1pt] (4.930,-1.050) -- (5.198,-1.050);
\fill[floorc] (5.049,-1.050) circle (0.075);
\draw[floorc,line width=1pt] (5.257,-1.470) -- (5.763,-1.470);
\fill[floorc] (5.495,-1.470) circle (0.075);
\draw[floorc,line width=1pt] (4.960,-1.890) -- (5.227,-1.890);
\fill[floorc] (5.079,-1.890) circle (0.075);
\draw[floorc,line width=1pt] (4.989,-2.310) -- (5.317,-2.310);
\fill[floorc] (5.138,-2.310) circle (0.075);
\draw[floorc,line width=1pt] (4.989,-2.730) -- (5.317,-2.730);
\fill[floorc] (5.138,-2.730) circle (0.075);
\draw[floorc,line width=1pt] (5.049,-3.150) -- (5.406,-3.150);
\fill[floorc] (5.198,-3.150) circle (0.075);
\draw[quantc,line width=1pt] (5.287,-3.790) -- (5.823,-3.790);
\fill[quantc] (5.455,-3.860) rectangle (5.595,-3.720);
\draw[quantc,line width=1pt] (6.001,-4.210) -- (6.835,-4.210);
\fill[quantc] (6.348,-4.280) rectangle (6.488,-4.140);
\draw[quantc,line width=1pt] (8.412,-4.630) -- (9.751,-4.630);
\fill[quantc] (8.967,-4.700) rectangle (9.107,-4.560);
\draw[quantc,line width=1pt] (8.382,-5.050) -- (9.692,-5.050);
\fill[quantc] (8.967,-5.120) rectangle (9.107,-4.980);
\node[font=\footnotesize\bfseries,anchor=south] at (12.900,0.100) {Clef (27B)};
\draw[gridc,line width=0.4pt] (10.400,-5.470) -- (10.400,0.000);
\node[font=\scriptsize,anchor=north,text=inkc] at (10.400,-5.500) {0\%};
\draw[gridc,line width=0.4pt] (11.829,-5.470) -- (11.829,0.000);
\node[font=\scriptsize,anchor=north,text=inkc] at (11.829,-5.500) {2\%};
\draw[gridc,line width=0.4pt] (13.257,-5.470) -- (13.257,0.000);
\node[font=\scriptsize,anchor=north,text=inkc] at (13.257,-5.500) {4\%};
\draw[gridc,line width=0.4pt] (14.686,-5.470) -- (14.686,0.000);
\node[font=\scriptsize,anchor=north,text=inkc] at (14.686,-5.500) {6\%};
\draw[inkc,dashed,line width=0.5pt] (11.055,-5.470) -- (11.055,0.000);
\draw[floorc,line width=1pt] (10.400,-0.210) -- (10.510,-0.210);
\fill[floorc] (10.400,-0.210) circle (0.075);
\draw[floorc,line width=1pt] (10.430,-0.630) -- (10.638,-0.630);
\fill[floorc] (10.519,-0.630) circle (0.075);
\draw[floorc,line width=1pt] (10.400,-1.050) -- (10.510,-1.050);
\fill[floorc] (10.400,-1.050) circle (0.075);
\draw[floorc,line width=1pt] (10.787,-1.470) -- (11.352,-1.470);
\fill[floorc] (11.055,-1.470) circle (0.075);
\draw[floorc,line width=1pt] (10.638,-1.890) -- (11.114,-1.890);
\fill[floorc] (10.876,-1.890) circle (0.075);
\draw[floorc,line width=1pt] (10.579,-2.310) -- (10.995,-2.310);
\fill[floorc] (10.787,-2.310) circle (0.075);
\draw[floorc,line width=1pt] (10.638,-2.730) -- (11.085,-2.730);
\fill[floorc] (10.846,-2.730) circle (0.075);
\draw[floorc,line width=1pt] (10.608,-3.150) -- (11.025,-3.150);
\fill[floorc] (10.817,-3.150) circle (0.075);
\draw[quantc,line width=1pt] (10.876,-3.790) -- (11.442,-3.790);
\fill[quantc] (11.074,-3.860) rectangle (11.214,-3.720);
\draw[quantc,line width=1pt] (11.204,-4.210) -- (11.918,-4.210);
\fill[quantc] (11.491,-4.280) rectangle (11.631,-4.140);
\draw[quantc,line width=1pt] (12.394,-4.630) -- (13.465,-4.630);
\fill[quantc] (12.830,-4.700) rectangle (12.970,-4.560);
\draw[quantc,line width=1pt] (13.823,-5.050) -- (15.102,-5.050);
\fill[quantc] (14.378,-5.120) rectangle (14.518,-4.980);
\fill[floorc] (1.200,-6.090) circle (0.075);
\node[font=\scriptsize,anchor=west] at (1.320,-6.090) {Not quantization};
\fill[quantc] (4.330,-6.160) rectangle (4.470,-6.020);
\node[font=\scriptsize,anchor=west] at (4.520,-6.090) {Quantization, within one engine};
\draw[inkc,dashed,line width=0.5pt] (9.600,-6.090) -- (10.100,-6.090);
\node[font=\scriptsize,anchor=west] at (10.150,-6.090) {Largest non-quantization source};
\end{tikzpicture}
\caption{Flip rates on the same 2,400 decisions, with 95\% intervals (exact Clopper-Pearson for zero-flip pairs). Each quantization row compares a build with BF16 in the same engine. The dashed line marks each model's largest non-quantization source.}
\end{figure}

At 4-bit the ordering holds on both models: every 4-bit effect exceeds every non-quantization source, with intervals that exclude zero (N2 below). At 8-bit it holds with three exceptions. Of the 32 differences between an 8-bit build and a non-quantization source across the two models, 29 exclude zero (exploratory; \nolinkurl{analysis/exploratory_checks.md}). The three that do not are Clef-flash\textquotesingle s Q8\_0 against its NVFP4 restart (21 flips against 20), and Clef\textquotesingle s Q8\_0 against its NVFP4 restart (25 against 22) and against its llama.cpp engine change (25 against 16). 4-bit is larger than 8-bit throughout. The 27B also differs from Clef-flash in two places. Its non-quantization sources other than the restarts are about twice as large: engine, batch and kernel changes flip 13 to 16 of its decisions, against 6 to 10 for Clef-flash. Its 4-bit GGUF build moves far less on this unit, 84 flips against 139.

The batch-size row is a rerun and a batch change at once. On Clef-flash we attribute its drift to the batch change, because the fallback path reproduces itself bit for bit across processes at batch 1, as the ANLI R1 check in the kernel section shows. Batching pads every sequence to the longest in its group, which changes the arithmetic, so a batch-size change is the plausible mechanism. That is an inference, not a measurement, and we did not repeat the batch-1 reproduction check on the 27B.

\subsection{4.2 Pre-registered tests}

We wrote five predictions on 2026-10-17, before analysing any quantization result, and report them as found. The tests name no model. \texttt{findings.md} scores them on Clef-flash, the model whose results existed when they were written, and those are the pre-registered outcomes. The 27B column applies the same tests to the second model and is exploratory.

{\def\LTcaptype{none}
\begin{longtable}[]{@{}
  >{\raggedright\arraybackslash}p{(\linewidth - 6\tabcolsep) * \real{0.0500}}
  >{\raggedright\arraybackslash}p{(\linewidth - 6\tabcolsep) * \real{0.3300}}
  >{\raggedright\arraybackslash}p{(\linewidth - 6\tabcolsep) * \real{0.3800}}
  >{\raggedright\arraybackslash}p{(\linewidth - 6\tabcolsep) * \real{0.2400}}@{}}
\toprule\noalign{}
\begin{minipage}[b]{\linewidth}\raggedright
\end{minipage} & \begin{minipage}[b]{\linewidth}\raggedright
Prediction
\end{minipage} & \begin{minipage}[b]{\linewidth}\raggedright
Clef-flash: test (difference of flip rates, pp {[}95\% CI{]}) and outcome
\end{minipage} & \begin{minipage}[b]{\linewidth}\raggedright
Clef, exploratory
\end{minipage} \\
\midrule\noalign{}
\endhead
\bottomrule\noalign{}
\endlastfoot
N1 & FP8\textquotesingle s flip rate against \texttt{bf16-srv} is not significantly larger than \texttt{bf16-srv} against its restart & +2.12 {[}+1.54, +2.71{]}. \textbf{Not supported} & +1.46 {[}+0.96, +1.96{]}. Not supported \\
N2 & NVFP4 and Q4\_K\_M exceed every non-quantization source, with intervals that exclude them & 16 differences, weakest +4.96 {[}+4.00, +5.96{]}; all exclude 0, also with Bonferroni (k = 16). Supported & 16 differences, weakest +2.58 {[}+1.79, +3.46{]}; all exclude 0, also with Bonferroni. Supported \\
N3 & Changing engine at BF16 moves at least as many decisions as 8-bit quantization within one engine & -1.79, -0.54, -1.88 and -0.62; every interval below 0. \textbf{Not supported} & -1.08, -0.50, -0.96 and -0.37; three intervals below 0, the fourth {[}-0.79, 0.00{]}. Not supported \\
N4 & Kernel selection flips a nonzero number of decisions, within the range of vLLM restart drift & 10 flips, against a restart range of 0 to 20. Supported & 14 flips, against a restart range of 0 to 22. Supported \\
N5 & llama.cpp at BF16 is bit-identical across restarts & 2,400 of 2,400 records. Supported & 2,400 of 2,400 records. Supported \\
\end{longtable}
}

\textbf{N1.} The prediction failed: FP8 changes 51 decisions where a BF16 restart changes none. The comparison was pre-registered against the BF16 restart. Against FP8\textquotesingle s own restart (5 flips), which was not pre-registered, the excess is still +1.92 pp {[}+1.33, +2.50{]}. The 8-bit GGUF build behaves the same way at a smaller size: Q8\_0 flips 21 against a llama.cpp floor of exactly 0, a difference of +0.88 pp {[}+0.54, +1.29{]}. The 27B repeats the pattern. Its FP8 build flips 39 against its own FP8 restart\textquotesingle s 0 (+1.62 pp {[}+1.12, +2.12{]}), and its Q8\_0 build flips 25 against a llama.cpp floor of 0 (+1.04 pp {[}+0.67, +1.46{]}).

\textbf{N3.} On both models, engine changes at BF16 are the smaller effect. On Clef-flash they flip 0.25\% to 0.33\% of decisions, against 0.88\% for Q8\_0 and 2.12\% for FP8. Our prediction overestimated the engine and underestimated 8-bit quantization. The gap is narrower on the 27B: its llama.cpp engine change (16 flips) and its Q8\_0 build (25) cannot be separated.

\textbf{N4.} The outcome depends on which restart the kernel switch is compared with. On Clef-flash its 10 flips lie above the BF16 vLLM restart (+0.42 pp {[}+0.21, +0.71{]}). They cannot be distinguished from the FP8 restart (+0.21 pp {[}-0.08, +0.54{]}) or the NVFP4 restart (-0.42 pp {[}-0.87, 0.00{]}). On the 27B the kernel switch (14 flips) lies above both its BF16 restart (+0.42 pp {[}+0.08, +0.75{]}) and its FP8 restart (+0.58 pp {[}+0.29, +0.88{]}), and cannot be distinguished from its NVFP4 restart (-0.33 pp {[}-0.83, +0.13{]}). The prediction as worded, nonzero and inside the restart range, holds on both.

\subsection{4.3 What the floor changes}

The floor is small next to every quantization effect we measured, and at 4-bit it is far from negligible.

A single-run NVFP4 comparison carries restart noise of up to 20 flips (Clef-flash) or 22 (Clef), against effects of 139 and 136, roughly a seventh. On both models the NVFP4 restart is the noisiest non-quantization source we found.

The vLLM restart floor depends on the format, but the two models disagree on how. Clef-flash\textquotesingle s floor rises with quantization: 0 flips at BF16, 5 at FP8 and 20 at NVFP4. Clef\textquotesingle s runs 4, 0 and 22. Its FP8 restart reproduced 95.3\% of records bit for bit, where Clef-flash\textquotesingle s FP8 restart reproduced 15.0\%. The share of bit-identical records does not track the flip count on either model. Clef-flash\textquotesingle s BF16 restart is 93.7\% bit-identical with 0 flips, while Clef\textquotesingle s is 6.7\% bit-identical with 4. The pattern we expected after the 9B, a floor that rises with fewer bits, does not survive the second model. What does survive is weaker and more useful. The floor is specific to the build and the model, so one noise number per engine is not enough, and each quantized build needs its own repeat. Our working hypothesis after the sweep was that kernel selection or autotuning at server start differs between instances. We tested it with a third instance, below, and it did not hold in its simple form.

Flips do not add across sources. On Clef-flash, FP8 against \texttt{ref} flips 47 decisions and FP8 against \texttt{bf16-srv} flips 51. On Clef the same two numbers are 33 and 39. (The counts against \texttt{ref} are in the "every config against ref" section of each model\textquotesingle s \texttt{decomposition.md}; the main table pairs builds inside one engine.) An engine change and a quantization step can cancel on some records and compound on others. A comparison that changes both at once reports neither effect correctly.

\textbf{A third server instance (exploratory, 2026-11-03).} After the sweep we started Clef\textquotesingle s FP8 and NVFP4 builds a third time (\texttt{fp8-r3}, \texttt{nvfp4-r3}), with the same flags, the same 2,400 records and the same 16 requests in flight. If the restart floor came from two discrete start-up states, such as a kernel configuration picked by timing at first launch (vllm\#58899), the third instance would match one of the first two record for record.

{\def\LTcaptype{none}
\begin{longtable}[]{@{}
  >{\raggedright\arraybackslash}p{(\linewidth - 6\tabcolsep) * \real{0.1500}}
  >{\raggedright\arraybackslash}p{(\linewidth - 6\tabcolsep) * \real{0.3600}}
  >{\raggedright\arraybackslash}p{(\linewidth - 6\tabcolsep) * \real{0.3600}}
  >{\raggedright\arraybackslash}p{(\linewidth - 6\tabcolsep) * \real{0.1300}}@{}}
\toprule\noalign{}
\begin{minipage}[b]{\linewidth}\raggedright
Pair of instances
\end{minipage} & \begin{minipage}[b]{\linewidth}\raggedright
FP8: flips {[}95\% CI{]}, bit-identical
\end{minipage} & \begin{minipage}[b]{\linewidth}\raggedright
NVFP4: flips {[}95\% CI{]}, bit-identical
\end{minipage} & \begin{minipage}[b]{\linewidth}\raggedright
NVFP4: mean TVD
\end{minipage} \\
\midrule\noalign{}
\endhead
\bottomrule\noalign{}
\endlastfoot
first vs second & 0, 0.00\% {[}exact 0.00, 0.15{]}, 95.3\% & 22, 0.92\% {[}0.54, 1.33{]}, 87.0\% & 0.00710 \\
first vs third & 0, 0.00\% {[}exact 0.00, 0.15{]}, 95.5\% & 22, 0.92\% {[}0.58, 1.33{]}, 78.6\% & 0.00754 \\
second vs third & 0, 0.00\% {[}exact 0.00, 0.15{]}, 96.5\% & 4, 0.17\% {[}0.04, 0.33{]}, 86.8\% & 0.00141 \\
\end{longtable}
}

A record counts as equal only if every question on it has bit-identical logits. The two-state prediction fails: the third instance differs from both earlier instances on 12.0\% of NVFP4 records (289 of 2,400) and 2.1\% of FP8 records (50 of 2,400). That rules out the two-state version of the kernel-selection story. A larger fixed set, or state that varies with timing, is not excluded by three runs.

The NVFP4 floor depends on run conditions, by more than five times: 4 flips between the second and third instances, 22 between either of them and the first. The first instance differs from the later two in two ways we cannot separate here: it compiled its graph from scratch where the later two loaded it from cache, and it scored the full task files rather than the subsets. The 4-flip pair shares the later runs\textquotesingle{} conditions (warm cache, subset streams) and the 22s compare across changed conditions, so 22 is the conservative bound for the effect-against-floor comparisons, and 4 to 22 is the range a replay can move depending on conditions. The first instance\textquotesingle s extra differences are spread across every position in each task: of the 224 NVFP4 records where the third instance equals the second but not the first, 120 fall in the first 100 records of their task, so they do not come from where the 200-record subsets end. Elapsed time is an unlikely third, since the second and third instances ran 12 days apart and still agree closely. FlashInfer\textquotesingle s FP4 kernel selection is probably not the cause, because the first instance tuned 110 kernel configurations and saved them, and the later two loaded those same 110.

The quantization effect barely moves. Against \texttt{ref}, the three FP8 instances flip 33, 33 and 33 decisions and the three NVFP4 instances 137, 139 and 137. A restart changes which near-tie decisions flip, and the count stays within two.

\subsection{4.4 Full-set effects}

All 31,275 public questions, each configuration against its own model\textquotesingle s \texttt{ref}. Accuracy and macro-F1 changes are in percentage points.

\textbf{Clef-flash (9B)}

{\def\LTcaptype{none}
\begin{longtable}[]{@{}lllllll@{}}
\toprule\noalign{}
Config & Accuracy & $\Delta$ accuracy {[}95\% CI{]} & $\Delta$ macro-F1 {[}95\% CI{]} & Agreement with \texttt{ref} & Brier & ECE \\
\midrule\noalign{}
\endhead
\bottomrule\noalign{}
\endlastfoot
\texttt{ref} & 82.2\% & & & & 0.2638 & 0.0355 \\
\texttt{bf16-srv} & 82.2\% & -0.01 {[}-0.06, +0.04{]} & -0.02 {[}-0.10, +0.06{]} & 99.8\% & 0.2641 & 0.0355 \\
\texttt{gguf-bf16} & 82.2\% & -0.01 {[}-0.05, +0.04{]} & -0.00 {[}-0.08, +0.08{]} & 99.8\% & 0.2641 & 0.0359 \\
\texttt{q8\_0} & 82.2\% & -0.00 {[}-0.08, +0.08{]} & -0.09 {[}-0.22, +0.04{]} & 99.4\% & 0.2643 & 0.0358 \\
\texttt{fp8} & 82.0\% & -0.20 {[}-0.32, -0.09{]} & -0.27 {[}-0.46, -0.08{]} & 98.8\% & 0.2658 & 0.0370 \\
\texttt{q4\_k\_m} & 81.9\% & -0.36 {[}-0.56, -0.18{]} & -0.64 {[}-0.95, -0.32{]} & 96.5\% & 0.2684 & 0.0343 \\
\texttt{nvfp4} & 81.3\% & -0.95 {[}-1.17, -0.71{]} & -1.55 {[}-1.90, -1.18{]} & 95.4\% & 0.2747 & 0.0366 \\
\end{longtable}
}

\textbf{Clef (27B)}

{\def\LTcaptype{none}
\begin{longtable}[]{@{}lllllll@{}}
\toprule\noalign{}
Config & Accuracy & $\Delta$ accuracy {[}95\% CI{]} & $\Delta$ macro-F1 {[}95\% CI{]} & Agreement with \texttt{ref} & Brier & ECE \\
\midrule\noalign{}
\endhead
\bottomrule\noalign{}
\endlastfoot
\texttt{ref} & 83.7\% & & & & 0.2344 & 0.0252 \\
\texttt{bf16-srv} & 83.7\% & -0.01 {[}-0.08, +0.06{]} & +0.01 {[}-0.10, +0.12{]} & 99.5\% & 0.2343 & 0.0251 \\
\texttt{gguf-bf16} & 83.7\% & -0.02 {[}-0.10, +0.05{]} & -0.02 {[}-0.12, +0.09{]} & 99.5\% & 0.2344 & 0.0255 \\
\texttt{q8\_0} & 83.6\% & -0.07 {[}-0.17, +0.02{]} & -0.10 {[}-0.25, +0.05{]} & 99.2\% & 0.2347 & 0.0255 \\
\texttt{fp8} & 83.6\% & -0.11 {[}-0.23, +0.02{]} & -0.10 {[}-0.29, +0.09{]} & 98.3\% & 0.2353 & 0.0261 \\
\texttt{q4\_k\_m} & 83.7\% & -0.02 {[}-0.20, +0.16{]} & +0.05 {[}-0.23, +0.32{]} & 96.8\% & 0.2372 & 0.0226 \\
\texttt{nvfp4} & 83.0\% & -0.65 {[}-0.87, -0.42{]} & -0.82 {[}-1.19, -0.50{]} & 95.2\% & 0.2443 & 0.0235 \\
\end{longtable}
}

Accuracy hides most of the movement. Clef-flash\textquotesingle s FP8 loses 0.20 points of accuracy while changing 374 decisions (1.20\%), and its Q8\_0 changes 183 decisions (0.59\%) with no measurable accuracy change at all. On the 27B, only NVFP4 loses measurable accuracy. Its Q4\_K\_M build changes 1,000 decisions (3.20\%) and its FP8 build 520 (1.66\%) while their accuracy stays inside an interval around zero. The changes go in both directions and mostly cancel in the aggregate. For any single user, though, they are real changes in the decision they receive.

The 27B\textquotesingle s full-set floor is also higher. On the full set, its engine changes flip 155 (vLLM) and 156 (llama.cpp) decisions and its batch-8 rerun 130, against 75, 67 and 66 for Clef-flash. Its 8-bit builds therefore sit at 1.7 to 4.0 times its own floor on the full set, and part of what reads as an 8-bit effect on the 27B is that floor.

\textbf{What the flips cost (exploratory).} A flip makes a decision worse (right to wrong against the gold label), better (wrong to right), or swaps one wrong answer for another. Accuracy counts only the difference between the first two. Splitting every full-set flip against \texttt{ref}:

{\def\LTcaptype{none}
\begin{longtable}[]{@{}lllllll@{}}
\toprule\noalign{}
Model & Build & Flips & Got worse & Got better & Wrong to wrong & Net change \\
\midrule\noalign{}
\endhead
\bottomrule\noalign{}
\endlastfoot
Clef-flash & Q8\_0 & 183 & 83 & 82 & 18 & -1 \\
Clef-flash & FP8 & 374 & 193 & 129 & 52 & -64 \\
Clef-flash & Q4\_K\_M & 1,108 & 527 & 413 & 168 & -114 \\
Clef-flash & NVFP4 & 1,426 & 769 & 472 & 185 & -297 \\
Clef & Q8\_0 & 264 & 126 & 104 & 34 & -22 \\
Clef & FP8 & 520 & 241 & 208 & 71 & -33 \\
Clef & Q4\_K\_M & 1,000 & 436 & 430 & 134 & -6 \\
Clef & NVFP4 & 1,502 & 762 & 560 & 180 & -202 \\
\end{longtable}
}

On Clef, Q4\_K\_M\textquotesingle s net change of six questions hid 436 that got worse. On Clef-flash the same build\textquotesingle s net change of 114 hid 527. Each build made between 83 and 769 decisions worse, while its net change ran from 1 to 297, so the net is the least informative of the four numbers. The split was not pre-registered, and it is computed by \nolinkurl{bench.exploratory_checks}.

\subsection{4.5 How far above the floor}

"Exceeds the floor" is a threshold. The ratio of each quantization effect to the largest non-quantization source on the same model and unit says by how much. The ratios are derived from the counts above (exploratory; \nolinkurl{analysis/exploratory_checks.md}).

{\def\LTcaptype{none}
\begin{longtable}[]{@{}lllllll@{}}
\toprule\noalign{}
Unit & Model & Floor used (flips) & Q8\_0 & FP8 & Q4\_K\_M & NVFP4 \\
\midrule\noalign{}
\endhead
\bottomrule\noalign{}
\endlastfoot
2,400 records & Clef-flash & vLLM NVFP4 restart (20) & 1.05 & 2.55 & 6.95 & 6.95 \\
2,400 records & Clef & vLLM NVFP4 restart (22) & 1.14 & 1.77 & 3.82 & 6.18 \\
31,275 questions & Clef-flash & engine change to vLLM (75) & 2.44 & 4.99 & 14.77 & 19.01 \\
31,275 questions & Clef & engine change to llama.cpp (156) & 1.69 & 3.33 & 6.41 & 9.63 \\
\end{longtable}
}

On the 2,400 records, Q8\_0 sits at the edge of the floor (1.05 and 1.14 times), FP8 1.8 to 2.6 times above it, and 4-bit 3.8 to 7 times. The denominator is a choice, and the reader should know four things about it. It is a different kind of source in each unit: a restart on the 2,400 records, and an engine change on the full set, where no restarts were run. The full-set numerators compare each build with \texttt{ref}, so they include an engine change as well. Against a build\textquotesingle s own restart instead, the llama.cpp ratios are undefined, because that floor is exactly zero, and Clef-flash\textquotesingle s FP8 is 10.2 (51 flips against 5). On Clef\textquotesingle s NVFP4 build, 22 is the largest of three restart pairs (4, 22 and 22). Two models from one family do not establish how the ratio changes with model size, and we do not extrapolate it.

\subsection{4.6 Where flips happen}

Flips concentrate where the reference is unsure. On Clef-flash, questions where the reference\textquotesingle s top probability is below 0.7 make up 16.3\% of the set. They hold 86\% of NVFP4\textquotesingle s flips, 90\% of Q4\_K\_M\textquotesingle s, 98\% of FP8\textquotesingle s and 100\% of Q8\_0\textquotesingle s. This is the pre-registered H3 result.

The highest reference confidence at which any Clef-flash build flipped a decision was:

{\def\LTcaptype{none}
\begin{longtable}[]{@{}ll@{}}
\toprule\noalign{}
Build & Highest \texttt{ref} confidence among its flips \\
\midrule\noalign{}
\endhead
\bottomrule\noalign{}
\endlastfoot
Q8\_0 & 0.663 \\
FP8 & 0.867 \\
Q4\_K\_M & 0.941 \\
NVFP4 & 0.969 \\
\end{longtable}
}

At 4-bit there is no clean "never above X" rule: Clef-flash\textquotesingle s NVFP4 flips 11 of the 18,125 questions where its reference is at least 90\% confident.

On the 27B (exploratory), 17.7\% of questions sit below 0.7. They hold 99\% of Q8\_0\textquotesingle s flips, 97\% of FP8\textquotesingle s, 89\% of Q4\_K\_M\textquotesingle s and 79\% of NVFP4\textquotesingle s. Neither 8-bit build flips any question where the 27B reference is at least 90\% confident. Q4\_K\_M flips 4 and NVFP4 28 of those 19,233 questions. The 27B\textquotesingle s NVFP4 build is the only build on either model with less than 80\% of its flips below 0.7. That is the bar the H3 rule sets, but H3 is scored on Clef-flash only.

\subsection{4.7 Two models in the same memory (exploratory)}

At 16.6 to 20.0 GB, the 27B\textquotesingle s two 4-bit builds sit within 15\% of Clef-flash\textquotesingle s BF16 builds (18.2 to 19.1 GB), the analyzer\textquotesingle s equal-memory window. On the same 31,275 questions, the 27B at Q4\_K\_M is 1.45 to 1.46 points more accurate than Clef-flash at BF16 on any of its three engines (for example +1.45 {[}+1.08, +1.86{]} against \texttt{ref}). At NVFP4 the lead is 0.82 to 0.84 points. Pre-registered H8 asked a different question: 4-bit Clef against 8-bit Clef-flash at equal memory. No such pair exists in this sweep, which the hypotheses section covers.

\section{5. Discussion}

The protocol we would ask of any quantization comparison follows directly from the measurements:

\begin{enumerate}
\def\labelenumi{\arabic{enumi}.}
\tightlist
\item
  \textbf{Compare inside one engine.} Serve BF16 on the same engine as the quantized build. On both models the engine change is smaller than the quantization effect, but it does not add to it in a simple way.
\item
  \textbf{Repeat the quantized build across a restart.} The floor is specific to the build and the model. It is zero for llama.cpp on both models, and 0 to 22 flips for vLLM, in an order that differs between the two models. Repeat it more than once and report the largest pair: on Clef\textquotesingle s NVFP4 build, three instances gave pairs of 4, 22 and 22 flips, so a single repeat can understate the floor by more than five times.
\item
  \textbf{Record the kernel that actually ran.} Pin or block optional kernel packages. Otherwise an unrelated environment change shifts results by about as much as a vLLM restart does.
\item
  \textbf{Report flips on paired records next to accuracy, and split them.} An accuracy change of -0.20 points was hiding 1.2\% of decisions changing on Clef-flash. On the 27B, an accuracy change indistinguishable from zero was hiding 3.2\%, and 436 of those decisions got worse. Accuracy is the decisions that got better minus the ones that got worse. Flips are both, plus wrong answers that changed to other wrong answers. Neither number is enough alone.
\item
  \textbf{Report the distance from the floor as a ratio, and name the denominator.} "Exceeds the floor" is a threshold, and an operator needs to know by how much.
\item
  \textbf{Keep nondeterminism and quantization apart.} A server that disagrees with itself has a defect to fix. A quantized build that disagrees with its BF16 original has a price to weigh. Calling both "changed decisions" merges the two.
\end{enumerate}

Most of this costs a few thousand extra inferences per configuration. Our repeats used 200 records per task, 2,400 in all, a small fraction of a full evaluation.

\textbf{Why llama.cpp is exactly reproducible here.} Its decision server processes one sequence at a time, so batch composition never varies. That is consistent with the zero floor on both models, but we have not verified the kernel-level reason and do not claim one.

\section{6. Limitations}

\begin{itemize}
\tightlist
\item
  \textbf{Two models, one family.} Both models use the same architecture (\nolinkurl{Qwen3_5ForConditionalGeneration}) and come from one release. The ordering of sources held on both. The vLLM restart pattern did not, and a third model could move it again.
\item
  \textbf{One device.} Every floor in this report was measured on one GB10, whose FP8 and NVFP4 kernels are young. Other hardware may show different restart behavior. Backend effects elsewhere can be much larger: Pape et al. (2026) report up to 16.6 points across five engines. We read our floor as a property of this device and these engines, not as a constant.
\item
  \textbf{Few restart replicates.} Each vLLM floor rests on one restart pair per build, with a third instance only for Clef\textquotesingle s FP8 and NVFP4 builds. That makes the floor the least certain term in every comparison against it. Replicates at five or more instances per build are the next measurement.
\item
  \textbf{Interval width.} 2,400 records give exact upper bounds of 0.15\% for zero-flip pairs. That is enough to separate every 4-bit effect from the floor, but not to rank the smaller non-quantization sources against each other, and on the 27B it cannot separate the llama.cpp engine change from Q8\_0.
\item
  \textbf{Batching inside vLLM runs.} Quality runs used 16 requests in flight, so each vLLM restart comparison includes differences in batch composition as well as instance differences. We report it as the restart drift a real deployment would see, not as instance drift in isolation.
\item
  \textbf{Restart mechanism open.} The third-instance probe rules out a two-state start-up explanation and makes FlashInfer\textquotesingle s FP4 kernel tuning an unlikely cause. It cannot separate a cold graph compile from the full-file versus subset request stream. Doing so takes a cold-cache instance on the subsets and a warm-cache instance on the full files.
\item
  \textbf{Confounded rerun.} \texttt{ref-r2} combines a rerun with a batch-size change. The attribution to batch size rests on the bit-for-bit reproduction at batch 1 on Clef-flash, and was not checked on the 27B.
\item
  \textbf{Pre-registered thresholds.} The decision rules for each hypothesis are given with its outcome in \texttt{analysis/findings.md}. Some carry judgment, such as what counts as "within the range" for N4, and we say where an outcome depends on it. For H2, the rule as written on 2026-10-15 did not say how to combine two 4-bit builds and two models. We added a reading on 2026-10-23, after the 27B results were seen, and it is dated as such.
\end{itemize}

\section{7. Pre-registered hypotheses}

Hypotheses H1 to H11 were registered on 2026-10-15 for the benchmark as a whole, before any result existed. Outcomes are as \texttt{findings.md} scores them. H2 and H8 involve both models; the rest are scored on Clef-flash.

H1 to H11 and the five tests N1 to N5 are separate registrations, two days apart. Two of them overlap. H1 asks N1\textquotesingle s question of both 8-bit builds, and H7 is N3. N2, N4 and N5 have no H counterpart. Where the two overlap, the pre-registered tests section gives the test as registered and this section scores the hypothesis as \texttt{findings.md} defines it.

{\def\LTcaptype{none}
\begin{longtable}[]{@{}
  >{\raggedright\arraybackslash}p{(\linewidth - 4\tabcolsep) * \real{0.0500}}
  >{\raggedright\arraybackslash}p{(\linewidth - 4\tabcolsep) * \real{0.4500}}
  >{\raggedright\arraybackslash}p{(\linewidth - 4\tabcolsep) * \real{0.5000}}@{}}
\toprule\noalign{}
\begin{minipage}[b]{\linewidth}\raggedright
\end{minipage} & \begin{minipage}[b]{\linewidth}\raggedright
Hypothesis
\end{minipage} & \begin{minipage}[b]{\linewidth}\raggedright
Outcome
\end{minipage} \\
\midrule\noalign{}
\endhead
\bottomrule\noalign{}
\endlastfoot
H1 & 8-bit is indistinguishable from BF16 within the noise floor & Not supported: FP8 and Q8\_0 both exceed their restart floor \\
H2 & 4-bit loses accuracy, more on Clef-flash than Clef in absolute flips & Mixed (see below) \\
H3 & Flips concentrate where ref confidence is low & Supported \\
H4 & Calibration (ECE/Brier) degrades before accuracy does & Not supported \\
H5 & Out-of-scope detection degrades more than in-scope classification & Mixed: FP8 only. On CLINC150 its out-of-scope recall falls 0.66 points {[}-1.33, -0.11{]} while in-scope accuracy holds. The 4-bit builds flip 2.0 to 2.6\% of out-of-scope items against 0.2\% of in-scope items, but in both directions, so recall does not fall measurably \\
H6 & Drift grows with input length, so real-world data drifts more than benchmarks & Mixed: grows with length for FP8 and both 4-bit builds (longest input quartile minus shortest: +0.35 points {[}+0.02, +0.69{]} for FP8, +1.15 and +1.20 for Q4\_K\_M and NVFP4); real-world data drifts more for all four builds, but low confidence explains that better than length does \\
H7 & Engine change moves decisions as much as 8-bit quantization & Not supported (same test as N3) \\
H8 & At equal memory, 4-bit Clef beats 8-bit Clef-flash on hard tasks but not easy ones & Not evaluable (see below) \\
H9 & Ordinal score questions shift toward the middle under 4-bit & Supported, and the shift is small: 0.013 to 0.129 of a scale point \\
H10 & A confidence-threshold cascade recovers near-BF16 agreement while escalating a small fraction & Mixed: 17 to 18\% escalation gives 99.5 to 99.7\% agreement on public tasks; the authors\textquotesingle{} private data needs 26 to 30\% \\
H11 & GGUF Q4\_K\_M and NVFP4 differ at the same nominal bit width & Supported: Q4\_K\_M is 0.59 points more accurate {[}0.37, 0.81{]} \\
\end{longtable}
}

\textbf{H2, mixed.} The rule is checked for each 4-bit build on the full public set, each model against its own BF16 reference. Three of the four builds lose accuracy (Clef-flash NVFP4 and Q4\_K\_M, Clef NVFP4). Clef\textquotesingle s Q4\_K\_M does not: -0.02 {[}-0.20, +0.16{]}. The second clause holds for Q4\_K\_M, where the 27B flips 1,000 decisions against Clef-flash\textquotesingle s 1,108. It fails for NVFP4, where the 27B flips 1,502 against 1,426. On the authors\textquotesingle{} private data the 27B flips fewer decisions for both 4-bit builds (917 against 2,425 for Q4\_K\_M, 1,195 against 2,168 for NVFP4). That comparison is reported beside the rule and is not part of it.

\textbf{H8, not evaluable.} H8 needs a 4-bit Clef build and an 8-bit Clef-flash build within the analyzer\textquotesingle s 15\% size window. The closest pair is Clef Q4\_K\_M (16.59 GB) against Clef-flash FP8 (12.15 GB), a ratio of 1.36. The comparison cannot be formed from the builds this sweep made, and no further run of this sweep changes that. Outside the window, and exploratory, the 27B\textquotesingle s 4-bit builds lead Clef-flash\textquotesingle s 8-bit builds by 5.0 to 6.4 points on the harder half of the tasks and trail by 0.6 to 1.2 points on the easier half. The split is a median split on Clef-flash\textquotesingle s reference accuracy, chosen after the 27B results were seen. We tested three alternative splits as a robustness check:

\begin{itemize}
\tightlist
\item
  a threshold at 80\% reference accuracy
\item
  a median split on the mean of both references
\item
  the median split with its two boundary tasks swapped
\end{itemize}

Under them the hard-task lead is 3.8 to 5.1 points and the easy-task deficit 0.4 to 1.5. The pattern holds in all twelve pair and split combinations. Much of the hard-task lead comes from one task: on RAGTruth alone the 27B Q4\_K\_M is 14.3 points ahead of Clef-flash FP8, while on SST-5 it is 0.5.

\section{8. Conclusion}

A quantization comparison on a production serving stack measures more than quantization. On two decision models and one GB10, sources that are not quantization moved up to 0.92\% of decisions: a vLLM restart up to 22 of 2,400, and a kernel chosen by which packages import 10 to 14. Every 4-bit effect cleared all of them, and 8-bit effects cleared them in 29 of 32 comparisons. Accuracy hid most of what quantization did, because the decisions it made worse and better largely cancelled. The checks that separate these effects cost a few thousand extra inferences per configuration: compare inside one engine, repeat each quantized build across a restart, record the kernel that ran, and report flips, split by direction, next to accuracy.

\section{Code and data availability}

Code, quantization recipes, build scripts, results and per-record outputs are at \url{https://github.com/Level5Tech/clef-quant-bench}. Quantized weights are at \url{https://huggingface.co/Level5Tech}.

Six tasks are released as records: BANKING77, CLINC150, MMLU, ARC-Challenge, BoolQ and When2Call. The other six (ANLI R1 to R3, ContractNLI, RAGTruth and SST-5) are released as record IDs, because their licenses restrict redistribution or do not clearly permit it. \texttt{datasets/build.py} rebuilds every record from pinned upstream revisions. We do not release the authors\textquotesingle{} private operational data or the project\textquotesingle s plan file, so the \texttt{-\/-private} analysis, \texttt{bench.private\_agg} and \texttt{bench.findings} below cannot be rerun, and \texttt{bench.speed} omits the private sets when rerun. Their aggregate outputs are in the repository, and the pre-registered text is in \texttt{PREREGISTRATION.md}.

Run from the repository root. The first command rebuilds the six tasks released as IDs and needs network access.

\begin{Verbatim}[breaklines=true,fontsize=\footnotesize]
python datasets/build.py
python -m bench.analyze --records data/records --results results --out analysis/flash_public --models clef-flash
python -m bench.analyze --records data/records --results results --out analysis/all_public
python -m bench.analyze --private private --out analysis/flash_private --models clef-flash
python -m bench.decompose --out analysis
python -m bench.decompose --model clef --out analysis/clef_decomp
python -m bench.speed
python -m bench.hypotheses
python -m bench.private_agg
python -m bench.findings --root . --out analysis/findings.md
python -m bench.restart_probes --records-dir data/records --results-dir results --model clef --out analysis_r3   # exploratory
python -m bench.exploratory_checks --records-dir data/records --results-dir results --out analysis   # exploratory
\end{Verbatim}

\section{Use of generative AI tools}

We used Claude (Anthropic), through Claude Code, throughout this work. It wrote the benchmark, sweep and analysis code to our specifications and operated the sweep on the GB10. It also searched the literature and model cards, and it drafted and edited this text, reference list included. An OpenCode agent reviewed earlier drafts for the second author. We fixed the report\textquotesingle s thesis and pre-registered hypotheses before analyzing any quantization result. The reported numbers come from the analysis scripts above. We reviewed the results and the text, and we take responsibility for all of it.

\section{References}
\begingroup\small\everypar{\hangindent=1.5em\hangafter=1}
Casanueva, I., Tem\v{c}inas, T., Gerz, D., Henderson, M., and Vuli\'c, I. (2020). Efficient Intent Detection with Dual Sentence Encoders. \href{https://arxiv.org/abs/2003.04807}{arXiv:2003.04807}.

Chen, T.-R. (2026). The Integer Alibi: Localizing Cross-Kernel Divergence in INT8-Quantized LLM Inference. \href{https://arxiv.org/abs/2608.13756}{arXiv:2608.13756}.

Clark, C., Lee, K., Chang, M.-W., Kwiatkowski, T., Collins, M., and Toutanova, K. (2019). BoolQ: Exploring the Surprising Difficulty of Natural Yes/No Questions. \href{https://arxiv.org/abs/1905.10044}{arXiv:1905.10044}.

Clark, P., Cowhey, I., Etzioni, O., Khot, T., Sabharwal, A., Schoenick, C., and Tafjord, O. (2018). Think you have Solved Question Answering? Try ARC, the AI2 Reasoning Challenge. \href{https://arxiv.org/abs/1803.05457}{arXiv:1803.05457}.

Clopper, C. J., and Pearson, E. S. (1934). The use of confidence or fiducial limits illustrated in the case of the binomial. Biometrika 26(4), 404-413. \url{doi:10.1093/biomet/26.4.404}.

Cloudflare (2026). Clef and Clef-flash model cards. \url{https://huggingface.co/Cloudflare/clef} and \url{https://huggingface.co/Cloudflare/clef-flash}.

Dutta, A., Krishnan, S., Kwatra, N., and Ramjee, R. (2024). Accuracy is Not All You Need. Advances in Neural Information Processing Systems 37. \href{https://arxiv.org/abs/2407.09141}{arXiv:2407.09141}.

fla-org. flash-linear-attention, version 0.5.2. \url{https://github.com/fla-org/flash-linear-attention}.

ggml-org. llama.cpp, build b11671. \url{https://github.com/ggml-org/llama.cpp}. Clef decision support added in pull request \#29831.

He, H., and Thinking Machines Lab (2025). Defeating Nondeterminism in LLM Inference. Thinking Machines Lab, 10 September 2025. \url{https://thinkingmachines.ai/blog/defeating-nondeterminism-in-llm-inference/}.

Hendrycks, D., Burns, C., Basart, S., Zou, A., Mazeika, M., Song, D., and Steinhardt, J. (2020). Measuring Massive Multitask Language Understanding. \href{https://arxiv.org/abs/2009.03300}{arXiv:2009.03300}.

Hugging Face. transformers, version 5.16.1. \url{https://github.com/huggingface/transformers}.

jpvelasco (2026). {[}Bug{]}: FlashInfer fp4\_gemm autotune changes temperature-0 tokens on Qwen3.8-27B NVFP4 (SM120). vllm-project/vllm issue \#62820, 21 October 2026. \url{https://github.com/vllm-project/vllm/issues/62820}.

Kasa, S. K., Kasa, S. R., and Negi, S. (2026). Same Text, Different Prediction: Serving-Context Nondeterminism in Text Classifiers. \href{https://arxiv.org/abs/2610.09111}{arXiv:2610.09111}.

Koreeda, Y., and Manning, C. D. (2021). ContractNLI: A Dataset for Document-level Natural Language Inference for Contracts. \href{https://arxiv.org/abs/2110.01799}{arXiv:2110.01799}.

Kozyrev, S., and Maiboroda, D. (2026). Why Gated DeltaNet Survives 4-Bit Quantization: NVFP4 W4A4 for the Recurrent Half of a Hybrid 27B LLM. \href{https://arxiv.org/abs/2609.04098}{arXiv:2609.04098}.

kurcontko (2026). clef-flash-FP8-Dynamic model card. \url{https://huggingface.co/kurcontko/clef-flash-FP8-Dynamic}.

Kwon, W., Li, Z., Zhuang, S., Sheng, Y., Zheng, L., Yu, C. H., Gonzalez, J. E., Zhang, H., and Stoica, I. (2023). Efficient Memory Management for Large Language Model Serving with PagedAttention. \href{https://arxiv.org/abs/2309.06180}{arXiv:2309.06180}.

Larson, S., Mahendran, A., Peper, J. J., Clarke, C., Lee, A., Hill, P., Kummerfeld, J. K., Leach, K., Laurenzano, M. A., Tang, L., and Mars, J. (2019). An Evaluation Dataset for Intent Classification and Out-of-Scope Prediction. \href{https://arxiv.org/abs/1909.02027}{arXiv:1909.02027}.

Li, B., Song, J., Fei, H., Hsu, W., and Lee, M. L. (2026). Flide: Accelerating Decision Inference with Exact State Forking. \href{https://arxiv.org/abs/2610.42148}{arXiv:2610.42148}.

Lu, D., and Ni, Z. (2026). How Far Does Second-Order Linear Response Predict Quantization Mismatch? Pre-Registered Measurements on Five Language Models. \href{https://arxiv.org/abs/2611.12634}{arXiv:2611.12634}.

Masoudian, S., Shafaei, P., Swain, M., and Schedl, M. (2026). What We Observe as LLM Behavior Can Be a Side-effect of Inference Backend. \href{https://arxiv.org/abs/2608.04714}{arXiv:2608.04714}.

Nie, Y., Williams, A., Dinan, E., Bansal, M., Weston, J., and Kiela, D. (2019). Adversarial NLI: A New Benchmark for Natural Language Understanding. \href{https://arxiv.org/abs/1910.14599}{arXiv:1910.14599}.

Niu, C., Wu, Y., Zhu, J., Xu, S., Shum, K., Zhong, R., Song, J., and Zhang, T. (2024). RAGTruth: A Hallucination Corpus for Developing Trustworthy Retrieval-Augmented Language Models. \href{https://arxiv.org/abs/2401.00396}{arXiv:2401.00396}.

Pape, D., Evertz, J., and Sch\"onherr, L. (2026). The Silent Hyperparameter: Quantifying the Impact of Inference Backends on LLM Reproducibility. \href{https://arxiv.org/abs/2605.19537}{arXiv:2605.19537}.

Patodiya, A. (2026). Same Request, Different Answer: Quantization Amplifies Cache-Induced Divergence in LLM Serving. \href{https://arxiv.org/abs/2609.04748}{arXiv:2609.04748}.

qgallouedec (2026). Warn once when a hub-kernel function falls back to its reference PyTorch path. huggingface/transformers pull request \#48185, merged 3 September 2026. \url{https://github.com/huggingface/transformers/pull/48185}.

Ross, H., Mahabaleshwarkar, A. S., and Suhara, Y. (2025). When2Call: When (not) to Call Tools. \href{https://arxiv.org/abs/2504.18851}{arXiv:2504.18851}.

Socher, R., Perelygin, A., Wu, J., Chuang, J., Manning, C. D., Ng, A., and Potts, C. (2013). Recursive Deep Models for Semantic Compositionality Over a Sentiment Treebank. Proceedings of EMNLP 2013, 1631-1642. \url{https://aclanthology.org/D13-1170/}.

TurinTech (2026). Clef-Artemis-GGUF model card. \url{https://huggingface.co/turintech/Clef-Artemis-GGUF}.

twu3202 (2026). {[}Bug{]}: Greedy output changes between restarts, also with VLLM\_BATCH\_INVARIANT=1: the q/k-norm + RoPE combo kernel picks its reduction config by timing. vllm-project/vllm issue \#58899, 27 September 2026. \url{https://github.com/vllm-project/vllm/issues/58899}.

vLLM project. LLM Compressor. \url{https://github.com/vllm-project/llm-compressor}.

Wu, Z., Dhiman, S., and Koshiyama, A. (2026). Which Decisions Low-Bit Quantization Breaks, and How to Predict Them. \href{https://arxiv.org/abs/2608.06564}{arXiv:2608.06564}.

Yang, S., Kautz, J., and Hatamizadeh, A. (2024). Gated Delta Networks: Improving Mamba2 with Delta Rule. \href{https://arxiv.org/abs/2412.06464}{arXiv:2412.06464}.

Yuan, J., Li, H., Ding, X., Xie, W., Li, Y.-J., Zhao, W., Wan, K., Shi, J., Hu, X., and Liu, Z. (2025). Understanding and Mitigating Numerical Sources of Nondeterminism in LLM Inference. \href{https://arxiv.org/abs/2506.09501}{arXiv:2506.09501}.
\par\endgroup

\section{\texorpdfstring{Appendix A. Per-task accuracy change against \texttt{ref} (full sets, percentage points {[}95\% CI{]})}{Appendix A. Per-task accuracy change against ref (full sets, percentage points {[}95\% CI{]})}}

\textbf{Clef-flash (9B)}

{\def\LTcaptype{none}\renewcommand{\tablesize}{\scriptsize}\setlength{\tabcolsep}{4pt}
\begin{longtable}[]{@{}llllll@{}}
\toprule\noalign{}
Task & \texttt{ref} accuracy & Q8\_0 & FP8 & Q4\_K\_M & NVFP4 \\
\midrule\noalign{}
\endhead
\bottomrule\noalign{}
\endlastfoot
ANLI R1 & 67.3\% & -0.20 {[}-0.80, +0.40{]} & -0.30 {[}-1.40, +0.70{]} & -1.70 {[}-3.40, -0.10{]} & -4.50 {[}-6.60, -2.60{]} \\
ANLI R2 & 58.3\% & -0.30 {[}-1.10, +0.40{]} & -0.50 {[}-1.60, +0.60{]} & -1.90 {[}-3.90, +0.00{]} & -2.20 {[}-4.10, -0.20{]} \\
ANLI R3 & 51.6\% & -1.08 {[}-1.83, -0.42{]} & -1.00 {[}-2.00, -0.08{]} & -1.67 {[}-3.50, +0.08{]} & -3.00 {[}-5.00, -1.08{]} \\
ARC-Challenge & 98.5\% & +0.00 {[}-0.26, +0.26{]} & +0.17 {[}-0.17, +0.51{]} & -0.09 {[}-0.43, +0.26{]} & -0.60 {[}-1.11, -0.09{]} \\
BANKING77 & 95.4\% & +0.00 {[}-0.10, +0.10{]} & -0.10 {[}-0.23, +0.03{]} & -0.06 {[}-0.32, +0.19{]} & -0.10 {[}-0.45, +0.26{]} \\
BoolQ & 91.1\% & +0.03 {[}-0.12, +0.18{]} & -0.18 {[}-0.52, +0.15{]} & -0.49 {[}-0.95, +0.03{]} & -0.49 {[}-1.07, +0.06{]} \\
CLINC150 & 98.0\% & +0.02 {[}-0.04, +0.10{]} & -0.14 {[}-0.26, -0.04{]} & +0.04 {[}-0.14, +0.24{]} & +0.06 {[}-0.16, +0.28{]} \\
ContractNLI & 74.1\% & -0.25 {[}-0.75, +0.25{]} & -0.45 {[}-1.10, +0.15{]} & +1.26 {[}+0.20, +2.31{]} & -0.30 {[}-1.41, +0.80{]} \\
MMLU & 91.5\% & +0.12 {[}-0.06, +0.32{]} & -0.20 {[}-0.46, +0.06{]} & -0.80 {[}-1.26, -0.34{]} & -1.18 {[}-1.70, -0.64{]} \\
RAGTruth & 67.4\% & +0.00 {[}-0.19, +0.19{]} & -0.19 {[}-0.44, +0.07{]} & -0.48 {[}-0.82, -0.15{]} & -0.93 {[}-1.37, -0.48{]} \\
SST-5 & 58.1\% & +0.32 {[}+0.00, +0.63{]} & -0.14 {[}-0.68, +0.41{]} & -0.45 {[}-1.40, +0.50{]} & -1.27 {[}-2.40, -0.23{]} \\
When2Call & 74.4\% & +0.19 {[}-0.16, +0.52{]} & -0.08 {[}-0.44, +0.33{]} & -0.08 {[}-0.79, +0.60{]} & -1.45 {[}-2.27, -0.66{]} \\
\end{longtable}
}

\textbf{Clef (27B)}

{\def\LTcaptype{none}\renewcommand{\tablesize}{\scriptsize}\setlength{\tabcolsep}{4pt}
\begin{longtable}[]{@{}llllll@{}}
\toprule\noalign{}
Task & \texttt{ref} accuracy & Q8\_0 & FP8 & Q4\_K\_M & NVFP4 \\
\midrule\noalign{}
\endhead
\bottomrule\noalign{}
\endlastfoot
ANLI R1 & 70.8\% & -0.60 {[}-1.60, +0.30{]} & +0.00 {[}-1.10, +1.10{]} & +0.00 {[}-1.40, +1.50{]} & -2.60 {[}-4.50, -0.80{]} \\
ANLI R2 & 64.1\% & -0.30 {[}-1.20, +0.60{]} & +0.40 {[}-0.80, +1.60{]} & +0.50 {[}-1.20, +2.10{]} & -2.30 {[}-4.30, -0.40{]} \\
ANLI R3 & 56.6\% & +0.17 {[}-0.58, +1.00{]} & -0.50 {[}-1.67, +0.67{]} & -0.33 {[}-1.92, +1.25{]} & -0.33 {[}-2.25, +1.67{]} \\
ARC-Challenge & 98.0\% & +0.00 {[}-0.26, +0.26{]} & -0.17 {[}-0.43, +0.00{]} & +0.09 {[}-0.34, +0.51{]} & -0.68 {[}-1.28, -0.17{]} \\
BANKING77 & 94.3\% & -0.16 {[}-0.39, +0.03{]} & -0.16 {[}-0.39, +0.06{]} & -0.06 {[}-0.45, +0.29{]} & -0.19 {[}-0.58, +0.19{]} \\
BoolQ & 91.2\% & +0.00 {[}-0.18, +0.18{]} & +0.06 {[}-0.21, +0.31{]} & +0.03 {[}-0.34, +0.40{]} & -0.18 {[}-0.64, +0.31{]} \\
CLINC150 & 97.3\% & -0.06 {[}-0.22, +0.10{]} & -0.06 {[}-0.24, +0.10{]} & -0.16 {[}-0.38, +0.08{]} & -0.14 {[}-0.40, +0.12{]} \\
ContractNLI & 77.3\% & -0.25 {[}-0.70, +0.15{]} & -0.95 {[}-1.66, -0.30{]} & +0.35 {[}-0.50, +1.16{]} & +0.00 {[}-1.10, +1.10{]} \\
MMLU & 89.7\% & +0.04 {[}-0.18, +0.26{]} & -0.14 {[}-0.46, +0.16{]} & -0.26 {[}-0.74, +0.22{]} & -1.38 {[}-1.94, -0.86{]} \\
RAGTruth & 81.1\% & -0.11 {[}-0.59, +0.33{]} & -0.44 {[}-1.07, +0.15{]} & +0.33 {[}-0.48, +1.11{]} & -1.41 {[}-2.52, -0.26{]} \\
SST-5 & 57.7\% & -0.05 {[}-0.45, +0.41{]} & +0.41 {[}-0.18, +1.04{]} & +0.68 {[}-0.18, +1.49{]} & +0.00 {[}-0.95, +1.04{]} \\
When2Call & 75.5\% & +0.00 {[}-0.22, +0.22{]} & +0.16 {[}-0.25, +0.55{]} & -0.47 {[}-1.04, +0.11{]} & -0.41 {[}-1.10, +0.30{]} \\
\end{longtable}
}

On Clef-flash the 4-bit losses concentrate on adversarial inference (ANLI) and knowledge (MMLU). On the 27B, only NVFP4 shows task-level losses that exclude zero: ANLI R1 and R2, MMLU, RAGTruth and ARC-Challenge. Intent classification is almost untouched at every bit width on both models.

One cell runs against the pattern. Clef-flash at Q4\_K\_M gains 1.26 points on ContractNLI {[}+0.20, +2.31{]}. It is the only task-level gain from a quantized build whose interval excludes zero; Clef-flash\textquotesingle s Q8\_0 on SST-5 (+0.32 {[}+0.00, +0.63{]}) touches it. The 27B\textquotesingle s Q4\_K\_M moves the same way on ContractNLI (+0.35 {[}-0.50, +1.16{]}). We have no explanation for it.

\section{Appendix B. Speed and memory on the GB10}

\subsection{B.1 Isolated latency}

One request at a time (batch 1, concurrency 1), 300 sampled records from each of the 16 sets (4,800 per configuration, 0 errors). Each configuration got 20 warm-up requests first. vLLM and llama.cpp time an HTTP round trip to localhost, including tokenisation on the server. The native reference times encode, collate, forward and synchronise in process. The public mean is the record-weighted mean over the 12 public sets. Peak memory comes from three sources:

\begin{itemize}
\tightlist
\item
  \nolinkurl{torch.cuda.max_memory_allocated} for the native reference
\item
  the vLLM plugin\textquotesingle s weights-plus-peak-activation figure for vLLM
\item
  \texttt{nvidia-smi} for the llama.cpp server process
\end{itemize}

Cross-engine memory figures are therefore approximate.

{\def\LTcaptype{none}\renewcommand{\tablesize}{\scriptsize}\setlength{\tabcolsep}{4pt}
\begin{longtable}[]{@{}
  >{\raggedright\arraybackslash}p{(\linewidth - 16\tabcolsep) * \real{0.1000}}
  >{\raggedright\arraybackslash}p{(\linewidth - 16\tabcolsep) * \real{0.1200}}
  >{\raggedright\arraybackslash}p{(\linewidth - 16\tabcolsep) * \real{0.1800}}
  >{\raggedright\arraybackslash}p{(\linewidth - 16\tabcolsep) * \real{0.0700}}
  >{\raggedright\arraybackslash}p{(\linewidth - 16\tabcolsep) * \real{0.0700}}
  >{\raggedright\arraybackslash}p{(\linewidth - 16\tabcolsep) * \real{0.1800}}
  >{\raggedright\arraybackslash}p{(\linewidth - 16\tabcolsep) * \real{0.0700}}
  >{\raggedright\arraybackslash}p{(\linewidth - 16\tabcolsep) * \real{0.0700}}
  >{\raggedright\arraybackslash}p{(\linewidth - 16\tabcolsep) * \real{0.0900}}@{}}
\toprule\noalign{}
\begin{minipage}[b]{\linewidth}\raggedright
Config
\end{minipage} & \begin{minipage}[b]{\linewidth}\raggedright
Engine
\end{minipage} & \begin{minipage}[b]{\linewidth}\raggedright
Clef-flash mean / p50 / p95 ms
\end{minipage} & \begin{minipage}[b]{\linewidth}\raggedright
Clef-flash public mean ms
\end{minipage} & \begin{minipage}[b]{\linewidth}\raggedright
Clef-flash peak GB
\end{minipage} & \begin{minipage}[b]{\linewidth}\raggedright
Clef mean / p50 / p95 ms
\end{minipage} & \begin{minipage}[b]{\linewidth}\raggedright
Clef public mean ms
\end{minipage} & \begin{minipage}[b]{\linewidth}\raggedright
Clef peak GB
\end{minipage} & \begin{minipage}[b]{\linewidth}\raggedright
Clef / Clef-flash (mean)
\end{minipage} \\
\midrule\noalign{}
\endhead
\bottomrule\noalign{}
\endlastfoot
\texttt{ref} & transformers & 354 / 172 / 1,577 & 381 & 19.97 & 1,162 / 594 / 5,005 & 1,243 & 56.27 & 3.28x \\
\texttt{bf16-srv} & vLLM & 184 / 105 / 704 & 196 & 19.93 & 596 / 348 / 2,170 & 635 & 55.90 & 3.24x \\
\texttt{fp8} & vLLM & 123 / 68 / 507 & 131 & 13.06 & 381 / 198 / 1,776 & 411 & 32.26 & 3.11x \\
\texttt{nvfp4} & vLLM & 88 / 50 / 352 & 93 & 10.00 & 270 / 148 / 1,048 & 284 & 21.75 & 3.08x \\
\texttt{gguf-bf16} & llama.cpp & 236 / 125 / 975 & 251 & 20.54 & 727 / 406 / 2,863 & 772 & 57.36 & 3.09x \\
\texttt{q8\_0} & llama.cpp & 255 / 127 / 1,053 & 267 & 12.95 & 799 / 457 / 3,259 & 846 & 33.16 & 3.13x \\
\texttt{q4\_k\_m} & llama.cpp & 254 / 121 / 1,121 & 267 & 9.52 & 802 / 414 / 3,512 & 848 & 21.65 & 3.16x \\
\end{longtable}
}

The 27B takes 3.1 to 3.3 times as long per decision as Clef-flash in every configuration. One request at a time, the engines split cleanly by format. Inside vLLM, FP8 is 1.5 times as fast as BF16 on both models and NVFP4 2.1 to 2.2 times. Inside llama.cpp, the quantized builds are 6 to 10\% slower than GGUF BF16. A decision is a single forward pass over the prompt with no generated text, so the workload is limited by arithmetic rather than by memory bandwidth, and smaller weights save less than they do for chat generation. The gain appears only where the engine runs the format on the GB10\textquotesingle s native low-precision arithmetic, as vLLM does for FP8 and NVFP4. This explanation is our working reading, and we have not profiled the kernels.

Latency grows with prompt length. CLINC150 and BANKING77 prompts list every intent, about 3,100 and 1,800 tokens on average. They are the slowest public sets in every configuration: 3.5 and 2.0 seconds per decision for the 27B Q4\_K\_M, against 0.27 seconds for SST-5.

Another process held GPU memory while every latency run started: 4.7 GB for most runs, and 10.4 GB for Clef-flash\textquotesingle s FP8, GGUF BF16 and Q4\_K\_M runs. This was not recorded for the native reference. The GB10\textquotesingle s memory is unified, so we do not expect a large effect, but we have not measured one.

\subsection{B.2 Throughput during quality runs}

Seconds per record over each configuration\textquotesingle s full quality run, model load excluded. A public record holds one question; a private record can hold several. These are throughput figures under each engine\textquotesingle s quality-run regime. vLLM had 16 requests in flight, llama.cpp ran one sequence at a time, and the native reference ran at batch 1 (\texttt{ref-r2} at batch 8). Cross-engine ratios therefore mix engine and batching effects. The Clef-flash column covers 12 sets (40,318 records) because four of its sets have no end line in the sweep log. The Clef column covers all 16 sets (44,690 records: 31,275 public and 13,415 private), so the two columns are not directly comparable. The pooled 27B/9B ratio, with per-set timing files filling the four gaps, is 3.14.

{\def\LTcaptype{none}
\begin{longtable}[]{@{}lll@{}}
\toprule\noalign{}
Config & Clef-flash s per record & Clef s per record \\
\midrule\noalign{}
\endhead
\bottomrule\noalign{}
\endlastfoot
\texttt{ref} (transformers, batch 1) & 0.472 & 1.384 \\
\texttt{ref-r2} (transformers, batch 8) & 0.664 & 1.887 \\
\texttt{bf16-srv} & 0.195 & 0.634 \\
\texttt{fp8} & 0.264 & 1.489 \\
\texttt{nvfp4} & 0.128 & 0.526 \\
\texttt{gguf-bf16} & 0.296 & 0.877 \\
\texttt{q8\_0} & 0.331 & 0.972 \\
\texttt{q4\_k\_m} & 0.336 & 1.024 \\
\end{longtable}
}

With 16 requests in flight, FP8 is slower than BF16 inside vLLM on both models, the reverse of the isolated latency result. The 27B FP8 build is the extreme case: 2.3 times as slow as the 27B\textquotesingle s vLLM BF16 overall, and 3.1 times as slow on CLINC150, the set with the longest prompts. We have not identified the cause. Batching at batch 8 made the PyTorch-fallback reference slower on both models, consistent with padding cost in a launch-bound path.

\section{Appendix C. Clef (27B) against Clef-flash (9B), both at BF16}

On all 12 public tasks (31,275 questions) the 27B reference is 1.47 points more accurate than Clef-flash\textquotesingle s: 83.69\% against 82.22\%, interval {[}+1.09, +1.85{]}. The largest gains are on RAGTruth (+13.74), ANLI R2 (+5.80), ANLI R3 (+5.00), ANLI R1 (+3.50) and ContractNLI (+3.26). It is behind on MMLU (-1.78), BANKING77 (-1.14) and CLINC150 (-0.70). Its top-1 agreement with the 9B reference runs from 68\% to 98\% by task and is 86.7\% pooled. That is below the lowest agreement of any 27B build with its own reference (95.2\%, NVFP4). The two models\textquotesingle{} disagreement is far larger than any quantization effect measured here.

\end{document}